\documentclass[conference]{IEEEtran}
\IEEEoverridecommandlockouts
\usepackage{cite}
\usepackage{amsmath,amssymb,amsfonts}
\usepackage{algorithmic}
\usepackage{graphicx}
\usepackage{textcomp}
\usepackage{xcolor}
\usepackage{hyperref}
\usepackage{tabularx}
\usepackage{booktabs}
\usepackage{orcidlink}
\usepackage{multirow}
\usepackage[table,xcdraw]{xcolor}

\makeatletter
\newcommand{\printfnsymbol}[1]{%
  \textsuperscript{\@fnsymbol{#1}}%
}
\makeatother

\def\BibTeX{{\rm B\kern-.05em{\sc i\kern-.025em b}\kern-.08em
    T\kern-.1667em\lower.7ex\hbox{E}\kern-.125emX}}
\begin{document}
\title{\textit{MeisenMeister: A Simple Two Stage Pipeline for Breast Cancer Classification on MRI}}
\author{
\IEEEauthorblockN{Benjamin Hamm\textsuperscript{1,2}\orcidlink{0009-0003-4818-8700},
Yannick Kirchhoff\textsuperscript{1,3,4}\orcidlink{0000-0001-8124-8435},
Maximilian Rokuss\textsuperscript{1,3}\orcidlink{0009-0004-4560-0760},
Klaus Maier-Hein\textsuperscript{1,2,3,4,5}\orcidlink{0000-0002-6626-2463}}
\IEEEauthorblockA{\textsuperscript{1}German Cancer Research Center (DKFZ) Heidelberg, Division of Medical Image Computing, Germany}
\IEEEauthorblockA{\textsuperscript{2}Medical Faculty, Heidelberg University, Germany}
\IEEEauthorblockA{\textsuperscript{3}Faculty of Mathematics and Computer Science, Heidelberg University}
\IEEEauthorblockA{\textsuperscript{4}HIDSS4Health, Karlsruhe/Heidelberg, Germany}
\IEEEauthorblockA{\textsuperscript{5}Pattern Analysis and Learning Group, Department of Radiation Oncology, Heidelberg University Hospital, Heidelberg, Germany}
\IEEEauthorblockA{\tt\small\{benjamin.hamm, yannick.kirchhoff, maximilian.rokuss\}@dkfz-heidelberg.de}
}

\maketitle
\section*{Introduction}

\noindent The ODELIA Breast MRI Challenge 2025 addresses a critical issue in breast cancer screening: improving early detection through more efficient and accurate interpretation of breast MRI scans. Even though methods for general-purpose whole-body lesion segmentation~\cite{lesionlocator} as well as multi–time-point analysis~\cite{longiseg} exist, breast cancer detection remains highly challenging, largely due to the limited availability of high-quality segmentation labels. Therefore, developing robust classification-based approaches is crucial for the future of early breast cancer detection, particularly in applications such as large-scale screening. In this write-up, we provide a comprehensive overview of our approach to the challenge. We begin by detailing the underlying concept and foundational assumptions that guided our work. We then describe the iterative development process, highlighting the key stages of experimentation, evaluation, and refinement that shaped the evolution of our solution. Finally, we present the reasoning and evidence that informed the design choices behind our final submission, with a focus on performance, robustness, and clinical relevance. We release our full implementation publicly at \url{https://github.com/MIC-DKFZ/MeisenMeister}

\section*{Materials and Methods}

\subsection*{Divide And Conquer Pipeline}

\begin{figure}[t]
  \centering
  \includegraphics[width=\linewidth]{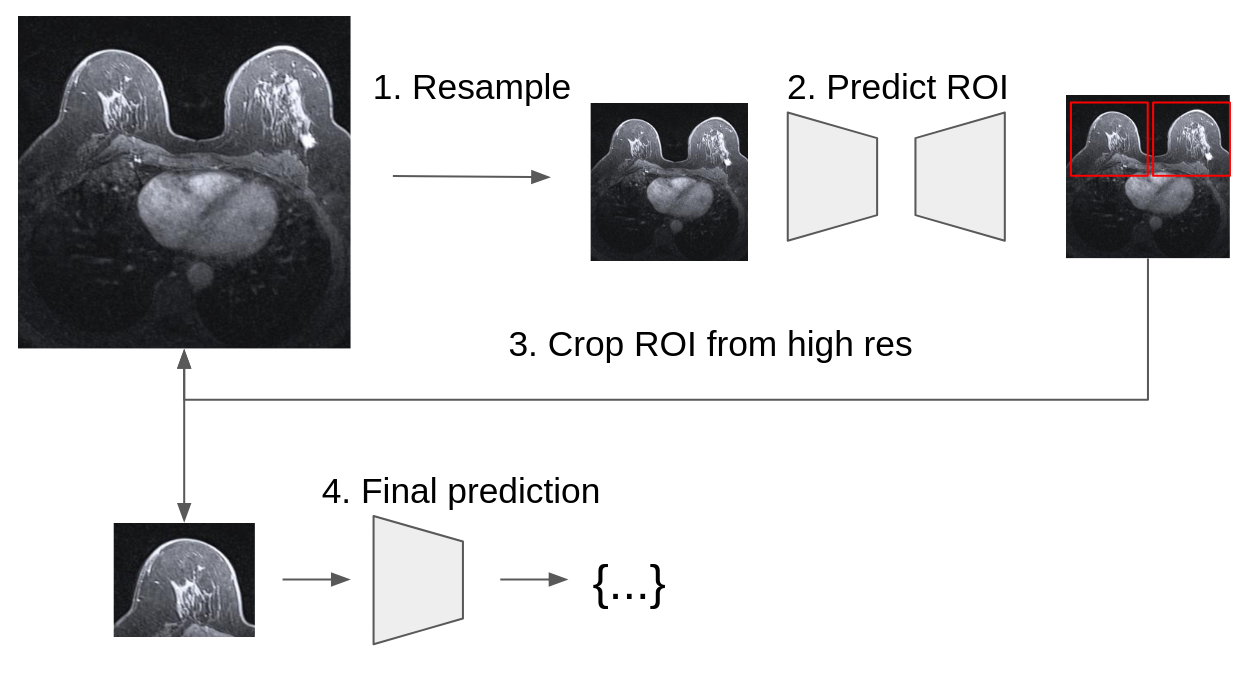}
  \caption{The pipeline follows a divide-and-conquer strategy to efficiently and accurately analyze high-resolution breast MRI volumes. (1) The original high-resolution 3D MRI is first resampled to a lower resolution to reduce computational complexity. (2) A segmentation model is applied to the resampled image to delineate the breast regions. Bounding boxes are then derived from the segmentation masks to identify the spatial extent of each breast. (3) Using these predicted bounding boxes, the corresponding high-resolution regions of interest (ROIs) are cropped directly from the original image, preserving anatomical detail. (4) Each high-resolution ROI is processed independently by a final prediction model, enabling focused, high-accuracy analysis at the level of individual breasts.}
  \vspace{-0.3cm}
  \label{fig:dac}
\end{figure}

\noindent We initiated our approach by designing a customized Divide-and-Conquer pipeline to address the computational challenges posed by high-resolution breast MRI volumes. Given the extremely high spatial resolution and large dimensionality of the input scans, processing an entire volume in a single forward pass is impractical due to memory limitations. Although the challenge organizers provided a unilateral cropping script to isolate a single breast per scan, our analysis showed that the resulting subvolumes remained too large for efficient model inference. To mitigate this, we developed a dedicated dataset and segmentation model, presented as \textit{Divide and Conquer: A Large-Scale Dataset and Model for Left–Right Breast MRI Segmentation}\cite{rokuss2025divideconquerlargescaledataset}. To improve performance and generalizability, we employed an active learning strategy that leveraged predictions from non–contrast-enhanced T1-weighted images as pseudo-ground truth for co-registered sequences, building on previously demonstrated reliability of cross-sequence annotation transfer~\cite{hamm2025enhancing}. This model enables accurate localization of individual breasts and the extraction of tight bounding boxes, thereby enabling efficient, ROI-focused processing in subsequent stages. A hierarchical overview of this pipeline is illustrated in Figure\ref{fig:dac}.

\subsection*{Public Data}

\noindent To enhance the challenge data, we sought to incorporate publicly available datasets. A primary challenge was identifying datasets with benign-labeled cases and harmonizing imaging modalities across varying acquisition protocols. We ultimately integrated two external datasets into our pipeline: the Advanced-MRI-Breast-Lesions (AMBL) dataset\cite{daniels2024advanced}, which includes both malignant and benign cases, and the Duke-Breast-Cancer-MRI (DUKE) dataset\cite{duke_breast_mri}, which contains only malignant cases. While DUKE does not include benign annotations, we utilized it in a separate experimental setting described later in the paper.

\subsection*{Model Training}

\noindent We utilized the nnU-Net framework \cite{isensee2021nnu} to create a classification training framework. Given the heterogeneity in available input modalities—particularly the variable number of post-contrast images and the optional presence of T2-weighted sequences—we conducted a series of ablation studies to determine the optimal input configuration. Specifically, we investigated (i) how many post-contrast phases were necessary for robust performance, and (ii) whether the inclusion of T2-weighted images contributed meaningful gains. \\

\noindent After finalizing the input channel configuration, we proceeded to ablate various backbone architectures for the classification model. Specifically, we compared three architectures: ResNet-18\cite{he2016deep}, ResEncL (a scaled-up variant of the nnU-Net encoder) and ResEncL with Squeeze-and-Excitation (SE) blocks\cite{hu2018squeeze}. A key consideration during this process was architectural compatibility with supervised segmentation-based pretraining. \\

\noindent We first conducted supervised pretraining on the MAMMA-MIA\cite{mama_mia_dataset} dataset using the standard nnU-Net framework. For this, we trained a single model (no ensembling) on the full dataset for 2000 epochs. While this prolonged training clearly led to overfitting, it served as a valuable initialization point for our downstream task. Following pretraining, we explored several fine-tuning strategies on the target classification task. Specifically, we compared: (i) linear probing, where only the classification head is trained; (ii) full fine-tuning of all model weights; and (iii) fine-tuning with learning rate warm-up, including two variants—one with a warm-up from 1e-4 to 1e-2, and another from 1e-5 to 1e-3. \\

\noindent After finalizing our fine-tuning strategy, we performed an ablation of data augmentation (DA) techniques. Effective augmentation was essential due to the limited number of training samples and the high susceptibility of models to overfitting in this setting. However, applying overly aggressive DA—particularly intensity-based transformations—can obscure diagnostically relevant features, effectively reducing the task to class distribution matching rather than promoting the learning of meaningful representations. To balance augmentation strength and fidelity, we systematically ablated individual DA components using only fold 0 (with CAM held out as the test center), as this fold showed good signal and resource constraints prevented full 5-fold ablation. The full list of compared DA techniques can be seen in Table\ref{tab:da_ablations}. These include contrast adjustment, gamma correction, Gaussian blur, Gaussian noise, multiplicative brightness, simulated low resolution, full 3D spatial transformations, in-plane spatial transformations, scaling, and elastic deformations. Each technique was tested in isolation to assess its effect on generalization and inform the final augmentation strategy used in our pipeline. In addition, we found that cropping away non-informative background regions by setting them to zero using the segmentation masks generated by our Divide-and-Conquer model led to significant performance gains.\\

\noindent Finally, we explored the potential of incorporating the DUKE dataset into our training pipeline, given its availability. However, DUKE contains only malignant cases, making it incompatible with the original multi-class classification task. To address this, we reformulated the problem as a binary classification task with two labels: healthy and lesion-present (i.e., benign + malignant). At inference time, we mapped the binary predictions back to the original three-class challenge setting. Specifically, if the model predicted lesion-present, we assigned its probability to the malignant class and attributed the remaining probability to benign. Conversely, if the model predicted healthy, we assigned that probability to the healthy class and the complement to benign. This approach inherently prevents the model from explicitly predicting the benign class. While this might appear limiting, we thought it could reduce confusion between healthy and benign, as well as between malignant and benign, ultimately improving overall class separation. Given that benign is a minority class in the dataset, we hoped this strategy could still yield a strong overall performance under macro-averaged metrics, despite the lack of a dedicated prediction path for benign cases. \\

\section*{Results and Discussion}

\noindent The results of our incremental pipeline improvements are summarized in Table\ref{tab:model_development}. We report the mean AUROC, averaged across five cross-validation folds, where each fold corresponds to holding out one acquisition center as the test set. Including the AMBL dataset provides a modest gain, while switching to the ResEncL backbone further improves performance. Fine-tuning strategies with learning rate warm-up offer clear advantages over linear probing, with warm-up schedules producing the largest single improvement—highlighting the impact of pretraining. Data augmentation and background masking contribute additional gains. The final configuration—combining ResEncL, warm-up fine-tuning, and data augmentation—achieves a strong baseline, which is further enhanced by the background masking step.

\begin{table}[ht]
\caption{Ablation results across key components of our pipeline. Reported AUROC values are averaged across 5 cross-validation folds, each corresponding to one held-out acquisition center. Bold entries indicate the setting we commit to use further.}
\centering
\begin{tabular}{@{\hspace{0.8ex}}l@{\hspace{0.8ex}}@{\hspace{2ex}}l@{\hspace{2ex}}@{\hspace{0.8ex}}c@{\hspace{0.8ex}}@{\hspace{0.8ex}}}
\toprule
\rule{0pt}{1.1EM}
\textbf{} &   & \textbf{Mean} \\
\textbf{Setting} & \textbf{Scheme} & \textbf{AUROC} \\ 
\toprule
\multirow{2}{*}{Data} 
        & Only Odelia Data & 0.623   \\ 
        & Add AMBL Data & \textbf{0.649}   \\ 
    \midrule
    \multirow{3}{*}{Networks} 
        & ResNet18 & 0.608  \\  
        & ResEncL SE & 0.631  \\ 
        & ResEncL & \textbf{0.649}  \\  
    \midrule
    \multirow{4}{*}{Finetuning (Mama Mia)} 
        & Linear Probing & 0.651   \\ 
        & Full Finetuning & 0.725  \\ 
        & Warmup 1e-4 1e-2 & 0.729   \\  
        & Warmup 1e-5 1e-3 & \textbf{0.738}  \\
    \midrule
    \multirow{1}{*}{Data Augmentation} 
        & Data Augmentation & \textbf{0.749}  \\ 
    \midrule
    \multirow{2}{*}{Preprocessing} 
        & Mask Background &  \textbf{0.765}   \\
    \cmidrule{2-3}
        & ResEncL + Warmup + DA & 0.736   \\ 
    \bottomrule
\end{tabular}
\label{tab:model_development}
\end{table}

\noindent The results of our data augmentation ablation study are summarized in Table\ref{tab:da_ablations}. Some techniques—particularly gamma transformation, elastic deformation, and simulated low resolution—led to performance degradation. These findings underscore the importance of carefully selecting augmentations in medical imaging tasks, where overly strong or unrealistic distortions can obscure clinically relevant features. The baseline model without augmentation achieves an AUROC of 0.826, which is exceeded by several well-calibrated augmentations, most notably Gaussian noise (0.851) and scaling (0.843).

\begin{table}[ht]
\caption{Ablation study of individual data augmentation techniques. Each augmentation was applied in isolation using the ResEncL backbone with warm-up fine-tuning. Reported AUROC values correspond to fold 0, where CAM was used as the held-out test center. Green cells indicate improvement over the no-augmentation baseline (0.826), while red cells indicate a performance drop.}
\centering
\begin{tabular}{l c}
\toprule
\textbf{Augmentation Technique} & \textbf{AUROC} \\
\midrule
Contrast Transform & 0.837 \cellcolor{green!25} \\
Gamma Transform & 0.823 \cellcolor{red!25} \\
Gaussian Blur Transform & 0.819 \cellcolor{red!25} \\
Gaussian Noise Transform & 0.851 \cellcolor{green!25} \\
Multiplicative Brightness Transform & 0.827 \cellcolor{green!25} \\
Simulate Low Resolution Transform & 0.812 \cellcolor{red!25} \\
Spatial Transform & 0.836 \cellcolor{green!25} \\
Spatial Transform Inplane & 0.813 \cellcolor{red!25} \\
Scaling & 0.843 \cellcolor{green!25} \\
Elastic Deform & 0.797 \cellcolor{red!25} \\
\midrule
No Augmentation & 0.826 \\
\bottomrule
\end{tabular}
\label{tab:da_ablations}
\end{table}

Batch size results are summarized in Table\ref{tab:bs}. Reducing the batch size increases gradient noise during training, which can serve as a beneficial regularizer. The results show a clear trend: smaller batch sizes consistently improve performance. A batch size of 1 yielded the highest mean AUROC of 0.765, averaged across five cross-validation folds. While training with very small batch sizes can introduce instability or slow convergence, in our setting it proved to be an effective and simple regularization strategy that improved generalization.

\begin{table}[ht]
\caption{Effect of batch size on classification performance using the ResEncL backbone with warm-up fine-tuning. Reported AUROC values are averaged across 5 cross-validation folds, each corresponding to one held-out acquisition center. Smaller batch sizes led to improved performance, with a batch size of 1 achieving the highest mean AUROC.}
\centering
\begin{tabular}{l c}
\toprule
 & \textbf{Mean} \\
\textbf{Batch Size} & \textbf{AUROC} \\
\midrule
4 & 0.745 \\
2 & 0.740 \\
\midrule
1 & \textbf{0,765} \\
\bottomrule
\end{tabular}
\label{tab:bs}
\end{table}

Given the variability in available MRI sequences across datasets, it was important to identify a modality combination that works for all centers. Including T2-weighted images alongside pre- and post-contrast sequences resulted in degraded performance (0.584), suggesting that the T2 modality either introduced noise or lacked sufficient cross-dataset consistency. Using only pre-contrast and two post-contrast phases (middle and last) significantly improved performance (0.636). The best results were achieved using the first and second post-contrast phases together with the pre-contrast scan (0.649), indicating that early dynamic enhancement patterns are particularly informative for classification. Results are shown in Table\ref{tab:channels},

\begin{table}[ht]
\caption{Effect of input channel configuration on classification performance. Reported AUROC values are averaged across 5 cross-validation folds, each corresponding to one held-out acquisition center. The combination of pre-contrast and early post-contrast phases yields the highest performance, while adding T2 leads to degradation.}
\centering
\begin{tabular}{l c}
\toprule
\textbf{Input Configuration} & \textbf{AUROC} \\
\midrule
Pre + Post middle + Post last + T2 & 0.584 \\
Pre + Post middle + Post last & 0.636 \\
\midrule
Pre + Post 1 + Post 2 & \textbf{0.649} \\
\bottomrule
\end{tabular}
\label{tab:channels}
\end{table}

To compare performance under different task formulations and preprocessing variants, we evaluated both the original three-class classification task ("Regular Task") and a simplified binary reformulation ("Binary Task") as described earlier. The results are summarized in Table\ref{tab:settings}. Each row incrementally adds components to the baseline configuration (ResEncL with warm-up fine-tuning), allowing us to isolate their effects. For the regular task, the best performance (0.776) was achieved with the inclusion of data augmentation, background masking and applying isotropic spacing. In contrast to our initial hypothesis, the binary task formulation did not consistently benefit from the incremental addition of components. Its best performance (0.757) was achieved when isotropic spacing was included, yet this still lagged behind the corresponding result in the regular task (0.776). These findings suggest that the trade-off between the number of benign cases and the confusion among the remaining classes did not yield the expected benefit.

\begin{table}[ht]
\caption{}
\centering
\begin{tabular}{@{\hspace{0.8ex}}l@{\hspace{0.8ex}}@{\hspace{2ex}}l@{\hspace{2ex}}@{\hspace{0.8ex}}c@{\hspace{0.8ex}}@{\hspace{0.8ex}}}
\toprule
\rule{0pt}{1.1EM}
\textbf{} &   & \textbf{Mean} \\
\textbf{Setting} & \textbf{Scheme} & \textbf{AUROC}   \\ 
\toprule
\multirow{4}{*}{Regular Task} 
        & ResEncL + Warmup &  \\  
        & + Data Augmentation & 0.765  \\   
        & + Mask Background &  \\ 
    \cmidrule{2-3}
        & + Isotropic Spacing & 0.776 \cellcolor{green!25} \\
    \midrule
    \multirow{4}{*}{Binary Task} 
        & ResEncL + Warmup & \cellcolor{red!25} \\  
        & + Data Augmentation & 0.722 \cellcolor{red!25} \\   
        & + Mask Background & \cellcolor{red!25} \\ 
    \cmidrule{2-3}
        & + Isotropic Spacing & 0.757 \cellcolor{red!25} \\
    \bottomrule
\end{tabular}
\label{tab:settings}
\end{table}

\section*{Conclusion}

Overall, we found it challenging to extract a stable and informative training signal from the available data, even after incorporating additional public datasets. Models exhibited a strong tendency to overfit, and training outcomes were highly sensitive to initialization and data splits, with substantial performance variance across runs using identical settings. This instability suggests a high level of noise in the learning process, making reproducibility difficult.

There remain several promising directions we have yet to explore. These include multi-resolution feature aggregation (e.g., in the spirit of HRNet\cite{wang2020deep}), additional regularization strategies to further combat overfitting, improved model selection techniques during training, and more extensive pretraining approaches. Additionally, gaining access to more data through secure federated learning\cite{mcmahan2017communication, hamm2025efficient} or by leveraging efficient annotation tools such as nnInteractive\cite{nninteractive} could be highly beneficial, as the challenge remains primarily a data limitation rather than a lack of strong architectural solutions.

\bibliographystyle{IEEEtran}
\bibliography{bibliography}
\end{document}